# Investigating the Ability of Deep Learning to Predict Welding Depth and Pore Volume in Hairpin Welding

Amena Darwish[*], Stefan Ericson, Rohollah Ghasemi, Tobias Andersson, Dan Lönn, Andreas Andersson Lassila, Kent Salomonsson

Virtual Manufacturing Processes, School of Engineering Sciences, University of Skövde, Kaplansgatan 11, Skövde 54134, Sweden

[*]Corresponding author: amena.darwish@his.se

## Abstract

To advance quality assurance in the welding process, this study presents a deep learning DL model that enables the prediction of two critical welds' Key Performance Characteristics (KPCs): welding depth and average pore volume. In the proposed approach, a wide range of laser welding Key Input Characteristics (KICs) is utilized, including welding beam geometries, welding feed rates, path repetitions for weld beam geometries, and bright light weld ratios for all paths, all of which were obtained from hairpin welding experiments. Two DL networks are employed with multiple hidden dense layers and linear activation functions to investigate the capabilities of deep neural networks in capturing the complex nonlinear relationships between the welding input and output variables (KPCs and KICs).

Applying DL networks to the small numerical experimental hairpin welding dataset has shown promising results, achieving Mean Absolute Error (MAE) values 0.1079 for predicting welding depth and 0.0641 for average pore volume. This, in turn, promises significant advantages in controlling welding outcomes, moving beyond the current trend of relying only on defect classification in weld monitoring, to capture the correlation between the weld parameters and weld geometries.

# 1  Introduction

Climate change is a serious global challenge, and the European Union's 2030 Climate Target Plan aims to substantially reduce at least 55% of greenhouse gas emissions compared to 1990 levels by 2030 [1]. Meeting the European Union's emission reduction goals heavily relies on electric powertrains. The key component in these powertrains is the electric motor, which needs to be highly efficient and high-performing. This requires an efficient winding and joining process for the copper pins in the motor parts, which increases power density. Traditional winding and joining technologies face limitations in terms of accessibility and possible damage to the insulating layer [2]. Therefore, implementing an innovative technology to minimize the power loss caused by the winding and joining process is important.

One effective approach to enhancing electric motor performance is called hairpin winding, which uses robust U-shaped insulated copper conductors with rectangular cross-sections [3]. In this process, these U-shaped conductors are inserted into the laminated core. The ends of the hairpins are then narrowed to create space for a twisting tool. This tool reshapes the hairpin ends so that matching ends of different pins are end-to-end. After twisting, the ends are joined, by laser welding, completing the hairpin winding. Finally, the stator undergoes impregnation and electrical testing [4] to ensure its functionality Figure 1.

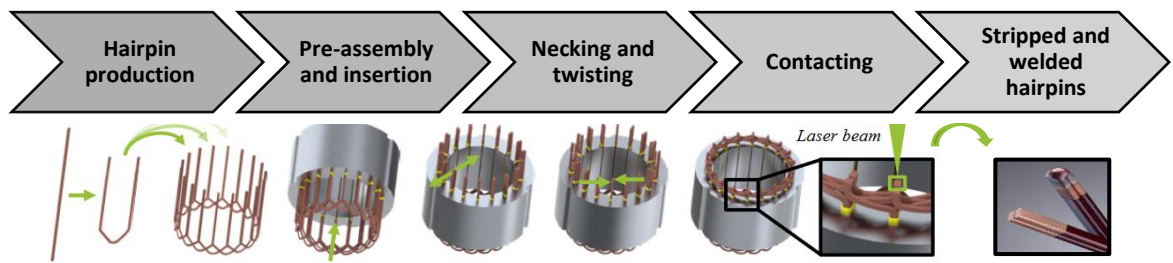

*Figure 1 The process chain for hairpin [5]*

Laser beam welding (LBW) is commonly recommended as a solution for hairpin winding due to its non-physical contact nature, high-speed beam, exceptional precision, energy density, automation capabilities, and superior weld quality [5-7]. Moreover, laser welding is highly versatile as it can be employed across a wide range of materials, including metals, alloys [1], and non-metallic materials [8, 9]. This characteristic of flexibility makes LBW widely applicable in different sectors, such as automotive [10], aerospace [11], medical device manufacturing [12], and other industrial applications [13].

Despite its unique advantages, laser welding is a complex manufacturing process due to its high energy density; the interaction between the laser beam and, in this case, the metal material is a complex physical phenomime where the metal is melting and solidification in a few milliseconds. The laser processes parameters such as welding speed and path, laser power, focused position, and the welding conditions, such as the microstructure of the materials and the cleanliness of the workpiece surfaces, significantly impact the quality of the output welds.

Simultaneously, the physical phenomena of melting and solidifying metal—such as keyhole formation, molten pool behavior, spatter, and plasma plume—significantly influence the weld output [7]. Therefore, there is a demand for in-situ laser monitoring approaches capable of observing the welding process to ensure high-quality welds.

Referring to the term "Monitoring" in this study, it encompasses more than just process observation; it includes data collection, laser process identification modeling, and defect detection [14]. Therefore, monitoring involves three stages of the welding process, each characterized by specific parameters and characteristics:

- The pre-processing stage: includes the laser beam parameters key input characteristics **(KICs)**. It is divided into changeable parameters (i.e. laser energy, the geometry of the beam path, and speed of the laser beam) and predefined parameters (i.e. the material properties, material thickness, workpiece conditions, and laser type). All **KICs** are set up before the welding process is started.
- The stage of in-process monitoring is dedicated to observing the welding zone, including Key Measurement Characteristics **(KMCs)**. It is about the physical phenomenon of melting and solidifying metal (i.e. keyhole formation, molten pool behavior, spatter, and plasma plume).
- Post-processing involves detecting and evaluating sample defects (i.e. pores, cracks, and keyhole collapse), and it is called Key Performance Characteristics (**KPCs**) [15]. The primary challenge in an in-situ monitoring case relates to the precise prediction and identification of the final welding characteristics of KPCs within a dynamic and nonlinear process distinguished by many interrelated factors, including KMCs and KICs. KPCs predictions rely significantly on computationally intensive multi-physics numerical simulations [16-23], making them inherently challenging and complex to solve. Nevertheless, an alternative approach instead of understanding the mechanism of the underlying phenomena [24] is using ML or DL as a part of ML (finding from data).

From a data acquisition and analysis perspective, the literature can be categorized by raw data type: image data and non-image data. The weld pool has a three-dimensional shape, but images of it are only two-dimensional. This means some information may be lost when viewing the weld pool in images. Skilled welders can see the weld pool in 3D, so images do not capture all the details they can see [25].

However, studies on deep learning-based welding image recognition (**DLBWIR**) are rapidly increasing due to the success of Convolutional Neural Networks (**CNNs**) in various applications. Another reason is that image data collection is easier compared to data collected from cutting, X-ray scans, microscopy, or simulations. Liu et al systematic review highlights the increasing number of publications employing DL to recognize weld images and monitor the welding process. Despite the rapid growth in research DL for welding image recognition, challenges in using image-based DL methods have also been identified [26]. Studies have documented that image-based DL methods are suitable for specific scenarios and vary from one application to another [25]. Additionally, CNNs are data-hungry methods, requiring the need to generate more data. Therefore, Generative Adversarial Networks (**GANs**) have been used to generate fake weld images to meet this data demand. [25,27,28]

Another type of raw data that are used in the mentoring methods is signal data from optical and/or acoustic sensors. With signal data, feature extraction is a key factor. Where extracting features and correlating them to the weld output KPCs is often a trial-and-error method. A further challenge is relying on weld images to understand signal behavior [14].

Numerical data is rare in the literature for several reasons. Collecting numerical data for each welding sample is expensive and time-consuming, which has limited the data collection efforts needed for DL models. However, there is a necessity for a DL model that can map the correlation between the weld input and weld output. Recent research has shifted towards building DL models as meta-models as a step to solve multi-objective Optimization problems, mapping the correlation between one of the KPCs and the selected input by studying it as an objective function. [29-31]. Or combine numerical models with DL to establish statistical or mathematical correlations with input-output data [32].

The novelty of this investigation arises from a gap in the literature, where predicting weld geometries and weld defects using machine learning from non-image data (numerical) is not a focal point of research. To highlight this gap, the literature is reviewed by categorizing it into two types: first, the use of ML to monitor and control KPCs from non-image data, and second, the application of DL to weld image datasets to predict one or more KPCs.

Yusof et al. [33] propose a DL model that utilizes an Artificial Neural Network **(ANN)**. Two KICs (Laser peak power and pulse duration) were used as input to the ANN. In addition to a set of statistical features extracted from an acoustic emission sensor. The model showed high accuracy in predicting weld depth and was trained using a dataset of 195 samples. Liu et al.'s model [34] utilized a hybrid methodology that combined the ANN with a genetic algorithm (GA) to predict both the number and average area of the porosities. In the Liu et al model, the inputs consist of four KICs variables (laser power, welding speed, focal position, and beam separation). Their model showed good accuracy and reduced the amount of porosities in welding 316L stainless steel. Luo and Shin [35] proposed a ML model for keyhole geometry formation in the laser welding process of Stainless steel 304. Their model used a radial basis function neural network that takes four KICs (laser power, welding speed, focal diameter, and approximate keyhole diameter) to predict the keyhole features (considering the penetration depth and inclination angle).

DL has shown more ability to capture complex attributes from the data, which could result in improved accuracy in classifying welding processes in comparison to the traditional method of manual feature engineering. Researchers are increasingly adopting DL techniques, mainly conventional neural networks **(CNNs)** [25], to analyze raw monitoring data from LBW image datasets [5] [36 - 42]. For instance, in Mayr et al.'s study [5], CNNs were applied to classify two-dimensional (2D) images obtained using a high-speed camera during hairpin welding. The classification process was conducted in both the pre-process and post-process stages, utilizing a dataset of about 500 images. The classification results demonstrated a high level of accuracy.

Similarly, in Zhao et al.'s model [37], an image processing technique was employed to examine various factors such as the coefficient of variation of vapor plume area **(VPA)**, vapor plume orientation **(VPO)**, and their correlation with the high-temperature zone **(HTZ)** of the molten pool that contains the keyhole, as well as the stability of the keyhole. In a study conducted by Hartung et al. [38], a novel approach was taken to develop a stacked dilated U-Net for in-process monitoring by adapting a vanilla U-Net model. They utilized high-speed camera imagery during the hairpin welding process to perform pixel-wise segmentation and accurately identify the spatter region. Shin et al. [39] used a DL model with CCD camera images as input to identify solidification cracks in AI 6000 alloy. Their model achieved 99.31% accuracy in prediction. Additionally, Knaak et al. [40] conducted a research study to investigate the effectiveness of various ML classification methods (KNN, SVM, MLP) as well as different CNN architectures (ResNet, Inception, MobileNetV2). The study aimed to classify six different defect classes and employed a substantial dataset of approximately 14,000 images for training and evaluation.

Meanwhile, addressing porosity issues in laser welding is critical, given the various sizes, shapes, and positions in which these defects are apparent [8], understanding the mechanisms responsible for pore formation is necessary. However, real-time monitoring of keyhole collapse to prevent subsequent events that may lead to pore formation proves even more demanding and is a highly challenging task [26]. Several factors could significantly affect the development process and formation of defects such as porosities and micro-cracks in the welding process. Hence, this requires a deep understanding and analysis of the keyhole's solidification phenomenon during welding to identify porosity regions and identify the cause of its formation.

Rivera et al. [24] utilized the random forest algorithm, SVM, and ANN to extract engineering features from weld images and analyze the correlation between these features and the porosities formation to classify weld quality in aluminum alloys. Ma et al. [36] utilized a Deep Belief Network **(DBN)** to classify the welding areas as either porosity or non-porosity regions based on the analysis of optical signals. However, due to its complexity, the prediction of pore volume was less accurate than the prediction of welding depth. In another attempt, Zhang et al. [42] demonstrated the capability of CNNs in detecting pores. Their findings revealed that CNNs achieved an impressive accuracy rate of 96.1%. However, they also mentioned that detecting deep and tiny pores presents a significant challenge.

Apart from having the previous studies in mind, this paper introduces a DL to predict two KPCs (welding depth and average volume of porosities) with a high accuracy using KICs (geometry, repetition of geometry paths, feed rates for each path, BLW ratio to the core for each path, and other parameters like machine energy output and nozzle usage during welding). The data are collected from numerical hairpin welding experiments, offering deeper insights into the relationship between KICs and the two selected KPCs. Unlike [5][36-42], this paper does not involve the classification of weld quality. Instead, it predicts numerical values for welding depth and average porosity volume using a range of KICs as input.

This research aims to develop models that enhance our understanding of the correlation between KPCs and KICs in hairpin welding by investigating DL capabilities. This study represents a key milestone in our ongoing research, which centers on implementing ML concepts to advance online quality assurance processes and techniques for laser welding applications.

The structure of the paper is as follows: section 2 clarifies the experimental preparations involving the preparation of Cu-ETP strips for hairpin welding and explains laser machine parameters. Subsequently, it explains the used dataset. In Section 3, the section proceeds to define the particulars of the designed DL model, followed by an evaluation of its performance. In Section 4, The results are thoroughly discussed and evaluated in detail.

# 2 Data preparation and explanation

## 2.1 Experiment setup

The present study investigates the effects of various laser welding parameters and welding geometry configurations on the joining of copper hairpins. Laser welding experiments are performed on specimens consisting of single hairpin pairs, considering different welding geometries, feed rates, and energy inputs. The specimens are made of pure enameled copper (Cu-ETP) wires with a height-to-width aspect ratio of approximately 0.65 and a cross-sectional area of approximately 8 $mm^2$. Before the experiments the specimens are cut to a length of 85 mm and the insulating enamel is mechanically removed over a length of 11 mm.

To reduce the gap between the individual hairpins, the specimens are clamped together using a customized fixture. Furthermore, to ensure a sufficient surface quality for the weld surface, and to reduce any height deviations between the individual hairpins, the top surface of the specimens is milled before the welding. The laser welding experiments are performed by the utilization of a Trumpf TruLaser Cell 3000 5-axis laser machine, equipped with programmable focusing optics (PFO) of the type PFO 33-2. The PFO is used for laser beam delivery and the realization of different welding geometries through remote laser welding.

The utilized laser source is a Trumpf TruDisk 6001, which is a 6kW solid-state disk laser. The generated laser light has a wavelength of 1030 nm and is delivered to the focusing optics through a 2-in-1 fiber, consisting of a core and ring fiber with diameters of 100 µm and 400 µm, respectively. The focal and collimation lengths for the focusing optics are 255 mm and 150 mm, respectively. This gives an aspect ratio of 1:1.7 between fiber and focal spot diameter. During the experiments, beam shaping is considered by utilizing the Bright Line Weld (**BLW**) technology developed by Trumpf [43].

With this beam-shaping technology, the power distribution in the focal point can be freely adjusted by controlling the total laser power allocated to the core fiber and the proportion allocated to the ring fiber. Using BLW in this experimental set-up allowed for expanding the keyhole's opening, helping release metal vapor and significantly reducing spatter occurrence [41].

## 2.2 Input features (KICs)

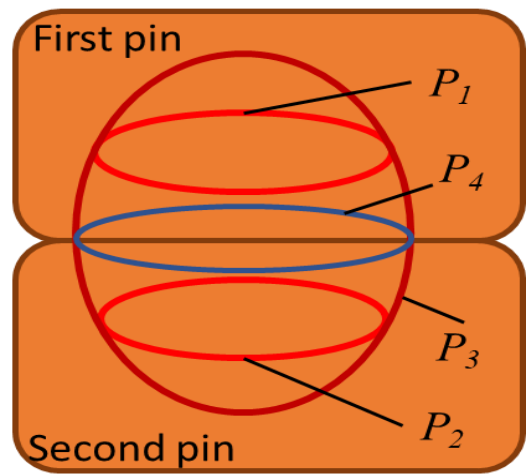

*Figure 2 General welding paths for the geometry.*

The data captured and used in the present study was obtained from a series of practical experiments conducted on 12 welding configurations consisting of 6 unique welding geometries according to Table 2. Each configuration involved three different welding paths, wherein a path represents the trajectory followed by the laser beam focal point moving across the workpieces. The current study also examines the various dimensions associated with these pathways. Figure 2 illustrates a general geometry with four welding paths with dimensions according to Table 2. The first path, $P_1$, consisted of multiple iterations arranged in an elliptical shape. The heat from $P_1$ affected the adjacent piece, increasing the temperature and heating the first pin. Subsequently, $P_2$ was designated as the second path, executed for the first hairpin, and required multiple repetitions. The molten regions of both workpieces fused, forming a mixed connection. $P_1$ and $P_2$ were conducted in opposing directions in certain samples.

The third step, denoted as $P_3$, aimed to increase the molten area's size and ensure a metal connection from the first and second pins. Unlike the elliptical shapes employed in $P_1$ and $P_2$, $P_3$ adopted a nearly circular shape with different repetitions. This choice of a circular shape was motivated by the intention to prevent the formation of "energy hotspots" in the gaps between the metal parts. Rapid movement of the laser beam focal point and pre-existing molten pools from $P_1$ and $P_2$ could lead to localized areas of high energy within the gaps, forming a hotspot [41]. In certain geometries, an additional path, denoted as $P_4$ was investigated in the neighborhood of the connection area. The used dataset contained categorical data of the geometrical features, while the geometrical dimensions were not incorporated as inputs in the model. Re-$P_1$, Re-$P_2$, Re-$P_3$, and Re-$P_4$ were quantitively represented as numerical data, representing the repetition of each path. The dataset includes information on the total length and duration of all paths for various geometries with different paths. The optimization of the feed rate is typically dependent on the specific material, thickness, and path of the welded joint. Other studies have shown that increasing the feed rate can stabilize the copper and aluminum welding process [11], [44].

For our experiments, the feed rates were also examined for each path at a laser power of up to 6 kW to optimize the welding process. Our dataset includes a feed rate feature for each path and the energy directed toward the core fiber of the laser machine. Due to the initial solid state of copper, a significant amount of energy reflection is expected. Therefore, a greater energy output would result in a more substantial heat input per unit of time. However, in $P_3$, excessive energy directed towards the core could lead to undesired melting or distortion of the workpiece. The feed rate varied from 150 mm/s to 500 mm/s across four paths. The laser's power intensity distribution in the core fiber and the ring differs between paths in different geometries. The dataset includes BLW ratio core % for paths ($P_1$-$P_3$), referred to as features $E_1$, $E_2$, and $E_3$. The range of all weld parameters is listed in Tables 1 and 2.

*Table 1 illustrates the welding parameters applied to the twelve geometries encompassed within the dataset (6 unique geometries and 12 configurations). Determining parameter variations is contingent upon the accumulated expertise and experiential knowledge in conjunction with the employed trial-and-error methodology during the study.*

| Welding parameter for 12 geometries | | | | |
| --- | --- | --- | --- | --- |
| **Designation of geometry** | $P_1$ | $P_2$ | $P_3$ | $P_4$ |
| **Number of geometry repetitions** | [2-4] | [1-2] | [1-3] | [0-2] |
| **Feed rate [mm/s]** | [180-321] | [180-321] | [150-500] | [0-290] |
| **BLW ratio core %** | [30-100] | [40-100] | [40-100] | 40 |

*Table 2. Dimensions of the welding paths for each unique welding geometry.*

| Geometry | Major radius | | | | Minor radius | | | |
| --- | --- | --- | --- | --- | --- | --- | --- | --- |
| | $P_1$ | $P_2$ | $P_3$ | $P_4$ | $P_1$ | $P_2$ | $P_3$ | $P_4$ |
| $G_1$ | 1.15 | 1.15 | 1.66 | - | 0.54 | 0.54 | 1.15 | - |
| $G_2$ | 1.15 | 1.15 | 1.66 | - | 0.27 | 0.27 | 1.15 | - |
| $G_3$ | 1.15 | 1.15 | 1.28 | - | 0.27 | 0.27 | 1.15 | - |
| $G_4$ | 0.99 | 0.99 | 1.28 | - | 0.27 | 0.27 | 0.84 | - |
| $G_4$ with an ellipse in the gap | 0.99 | 0.99 | 1.28 | 1.28 | 0.27 | 0.27 | 0.84 | 0.34 |
| $G_4$ with a line in the gap | 0.99 | 0.99 | 1.28 | 1.28 | 0.27 | 0.27 | 0.84 | 0.00 |

## 2.3 KPCs (model output)

In our used dataset, two KPCs were featured as outputs: Welding depth [mm] and average porosity volume [mm$^3$]. These measurements were obtained using an RX solution computed tomography (CT) scanner. In addition to that, the welding depth was further confirmed in specific samples through a combined metallographic examination. For the CT scan examinations, the welded samples underwent scanning in the CT scanner using a voxel size of 50 µm.

The minimum porosity volume was identified as 2×2×2 voxels size. A CT scan was acquired for each welding joint. The average pore volume [mm$^3$] was measured using "VG Studio Max." The software mentioned above was utilized to measure the depth of welding manually. Figure 3 shows an example of a measurement of pore volume and welding depth from the CT scan.

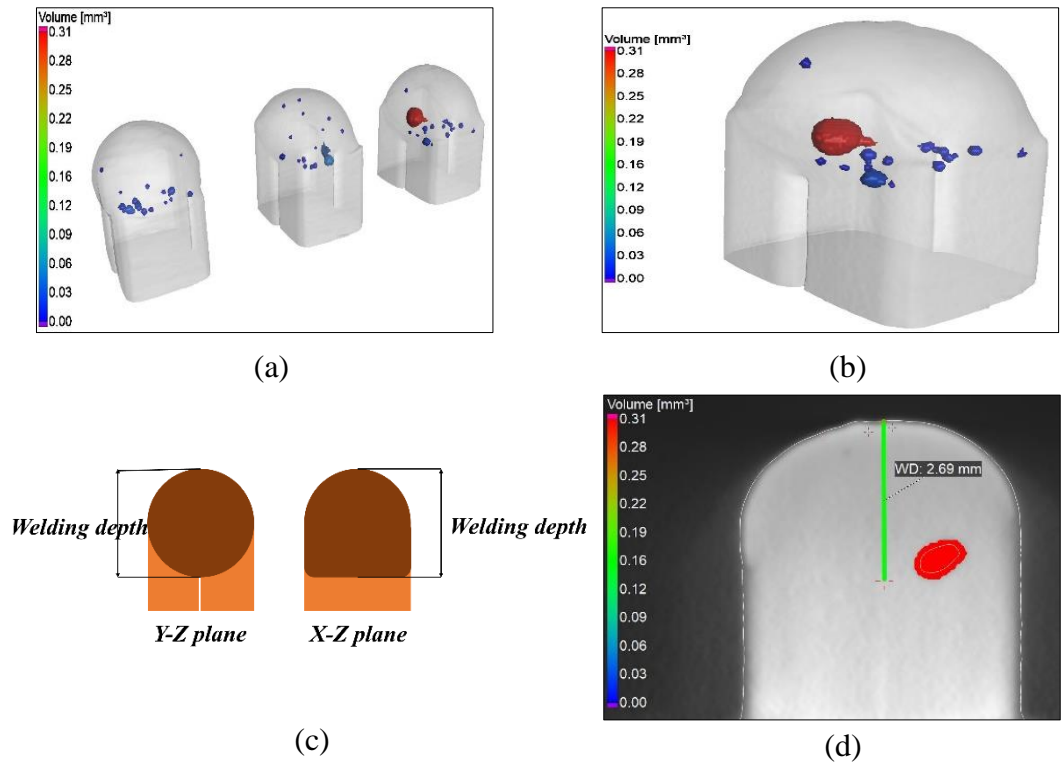

*Figure 3 illustrates porosities and welding depth measurements in CT scans Using VG Studio Max Software. (a) volume of porosities for three hairpin samples, with a color bar representing varying porosity volumes, (b) porosity volume within one of the three samples, (c) visual representations of welding depth in hairpins from two perspectives (Y-Z and X-Z), (d) the manual measurement of welding depth using the software, achieved by selecting the endpoint of the depth based on the small gap observed between the two hairpins in the Y-Z view.*

## 2.4 Exploratory Data Analysis

After conducting experiments and measuring the (KPCs), the final dataset contained 134 samples. Geometries are stored as categorical data, as shown in Table 2. For mathematical computations in DL, numerical representation is needed. Therefore, during the preprocessing stage, label encoding is used to convert categorical data into numerical form. Additionally, to ensure all data points contribute equally, normalization is required. In our case, since the data values have a narrow range (for example, weld depth values differing by fractions of a millimeter), normalization improves the variability between data points. To achieve this MinMaxScaler from the Sklearn library in Python has been used. For every feature, the minimum value of that feature gets to 0, the maximum value gets to 1, and every other value gets transformed into a decimal between 0 and 1 following this equation:

$$\text{Current value} = \frac{Value - min}{\max - min}$$

This improves the model accuracy because every data point has the same scale so each feature is equally important.

To reduce the input dimensions this study followed [45, 46] by Investigating the correlation between the KICs and KPCs. Sinha [45] utilized statistical correlations before applying ML techniques to explain the relationship between welding parameters. Also reducing the number of input features for the DL model through correlation analysis. The presented correlation in this research indicates the degree of linear interdependence between variables,

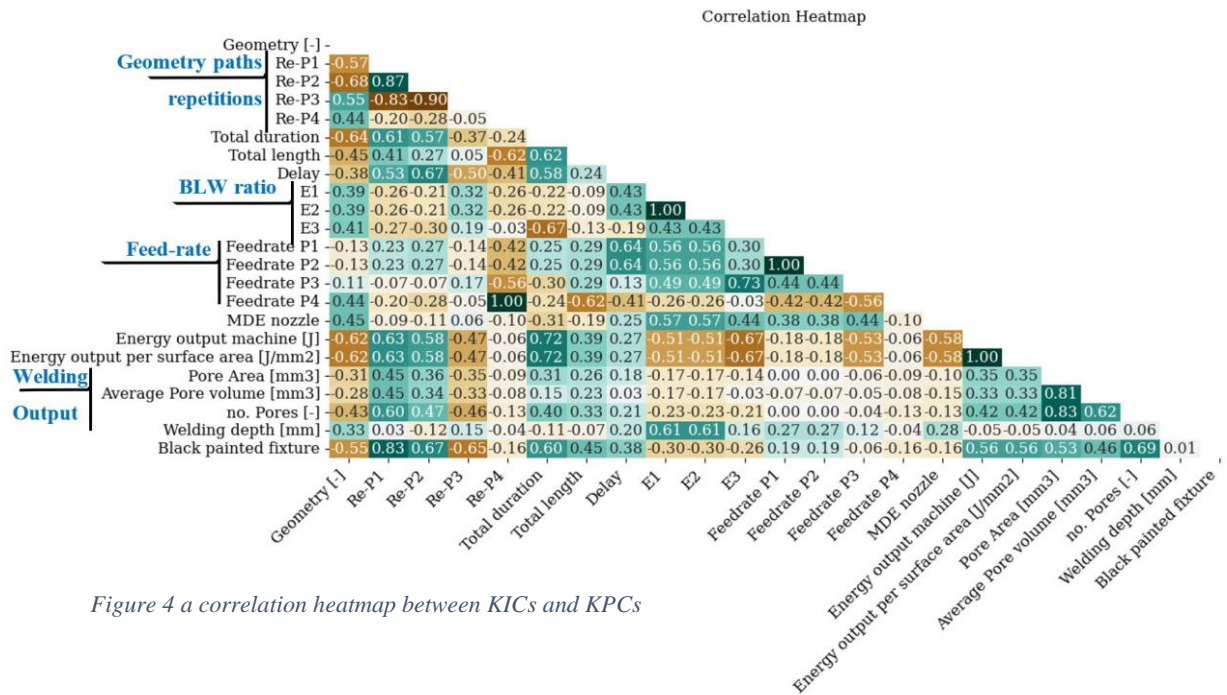

*Figure 4 a correlation heatmap between KICs and KPCs*

which has been determined by calculating covariance. To evaluate the relationship between variables, the correlation is considered strong if ($|Corr| > 0.7$), or weak ($|Corr|$ 0.4) in a range from −1 to 1.

Two KPCs are chosen as output (welding depth and average pore volume). The results were presented in Python using Seaborn heatmaps, as illustrated in Figure 4. The heatmap correlation showed a significant correlation between ($E_1$, and $E_2$) which are the variables that indicate the BLW ratio for the first and second paths in the weld geometry. Also, $E_1$ and $E_2$ have a strong correlation to the weld depth. Moreover, a weak correlation was observed between the number of repetitions for the first path and the welding depth. This could be attributed to the welding depth not being significantly influenced or governed during this initial material heating stage [7].

Additionally, a strong correlation between the feed rate for the first and second paths and the weld depth. Furthermore, it was noticeable that Re-$P_4$ did not significantly affect the welding depth. Furthermore, a strong positive correlation was observed between the pore volume/number of pores and path repetitions. It was also noted that an increase in Re-$P_3$ led to decreased pore volume/number of pores. And that is why $P_3$ and $P_4$ are used, to reduce the number of pores and increase the weld quality.

## 2.5  Distribution

For each set of input parameters, we have five repetitions, meaning that for the same input, the dataset has five different output values. This variability affects the neural network during training if the output values show strong variations. Therefore, there is a need to visualize and understand the level of stochasticity in the output variables. This analysis aims to show whether the output variables (weld depth and pore volume) show stochastic behavior.

Figure 5 (a) demonstrates that the welding depth variable follows a normal distribution. A variable following a normal distribution has a well-behaved linear model, which means a higher accuracy in the DL model. Figure 5 (b) illustrates the distribution of the pore volume within the dataset and the degree of deviation from the normal distribution. To address the randomness, we calculate the average of these repetitions for each set of input parameters. However, it is important to note that the pore volume variable demonstrates randomness in the data.

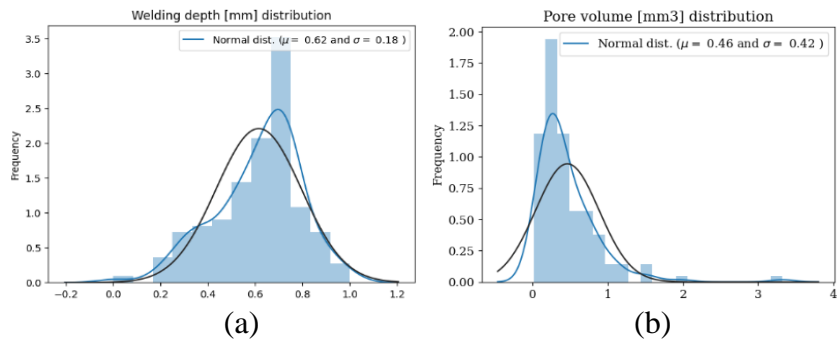

*Figure 5 displays the distribution of the output variables the black line on the histogram o represents the output data variables. (a) illustrates the distribution of the first output variable (welding depth), (b) the distribution of the second output (pore volume).*

# 3 Methodology

## 3.1 Design 2 DL networks to estimate the two of the KPCs (welding depth and average pore volume)

As mentioned in the introduction, DL has become a powerful tool for modeling complex processes and making accurate predictions. This is due to its ability to learn and extract hidden patterns from datasets and capture non-linear relationships between variables. This study utilizes a multilayer feed-forward neural network, where data flows from the input layer through several hidden layers to the output layer, with each layer's output serving as the input for the next layer.

The DL model was trained using a back-propagation algorithm. In this algorithm, predictions are compared to actual values to compute an error, which is then propagated back through the network to adjust the weights. Each neuron in the network has parameters called weights, which are modified during the learning process. These adjustments occur iteratively to minimize the error over time.

The difference between predicted values and actual values is measured using a loss function. In this study, the Mean Squared Logarithmic Error **(MSLE)** is used as the loss function. MSLE is selected because it is suitable for non-negative target variables and reduces the impact of outliers, thereby ensuring more balanced percentage errors.

$$MSLE = \frac{1}{n}\sum_{i=1}^{n}(log(1 + predicted\ output_i) - log(1 + actual\ output_i))^2 \qquad (1)$$

The dataset consists of 20 features, and there is a total of 134 samples available for analysis. The dataset was divided into 80% for training, 10% for validation, and 10% for testing.

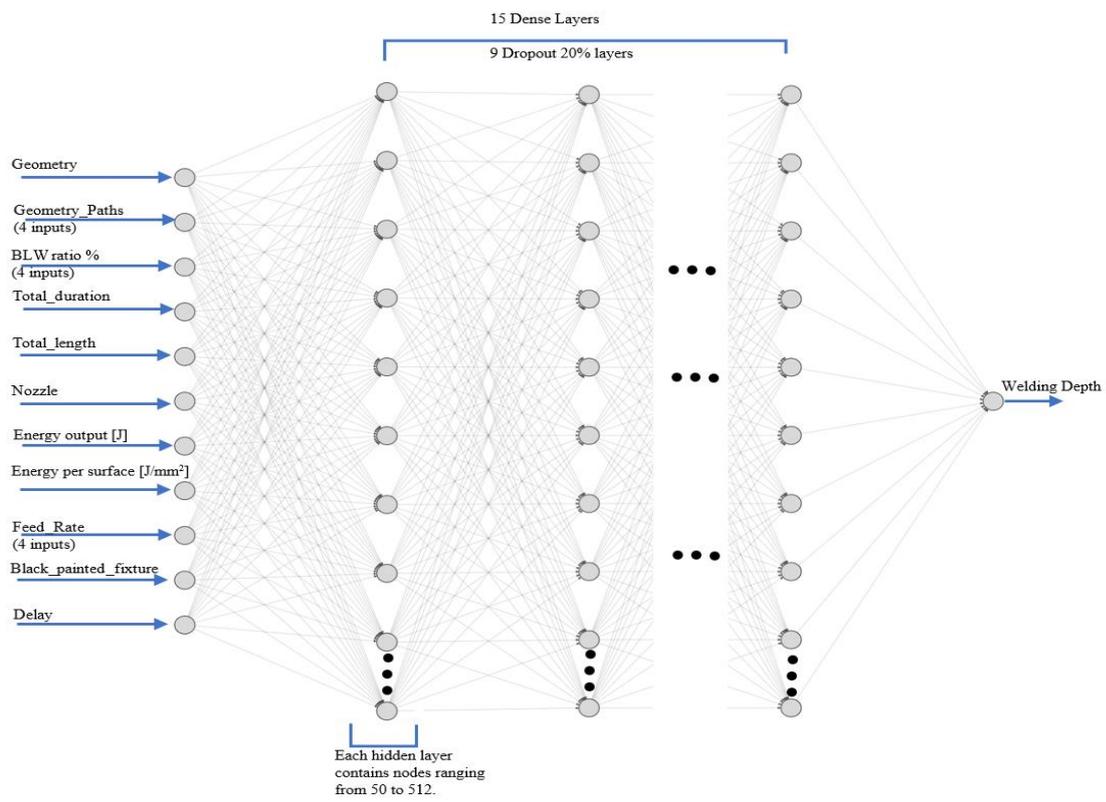

*Figure 6 deep learning architecture for welding depth prediction.*

To evaluate the accuracy of the model, three statistical metrics were computed (Equations 1, 2, and 3).

First, the loss function was calculated for both the training and validation epochs. Second, the Mean Absolute Error **(MAE)** was used as a statistical metric to assess the model's performance. MAE has a similar behavior towards outliers, making it suitable for this study with the chosen loss function. It quantifies the average absolute difference between the predicted and actual values.

$$MAE = \frac{\sum_{i=1}^{n}|predicted\ output_i - actual\ output_i|}{n} \qquad (2)$$

A third measure is needed because the model is learning from a small dataset. It's important to remember that "correlation does not imply causation." This means that while correlations can sometimes suggest possible relationships between variables, a positive correlation between two variables doesn't prove that changing one variable directly causes changes in the other.

The coefficient of determination **($R^2$)** is used to measure the goodness of fit of a model. $R^2$ is a statistical measurement to represent the average change in the output variable for each unit change in the input variables. Including $R^2$ as an evaluation metric enables examining the model's capacity beyond the training dataset.

$$R^2 = 1 - \frac{\sum_{i=1}^{n}(actual\ value - predicted\ value)^2}{\sum_{i=1}^{n}(actual\ value - mean\ value)^2} \qquad (3)$$

Neurons in the input layer do not use any activation function, while neurons in the hidden layers employ a Rectified Linear Unit **(ReLU)** activation function. This choice of activation function allows the neurons to produce positive outputs, which helps the network learn faster. Neurons in the output layer employ a linear activation function, allowing the predicted values to be continuous and take any real number, which is suitable for regression tasks.

To adjust the DL parameters during the training phase, we used Adam as the optimizer function. This choice handles sparse gradients (where gradients are zero) and adjusts learning rates based on recent gradient magnitudes. Adam optimizer uses the default learning rate of 0.01. Following this, a model tuning process was executed, involving the exploration of various parameters, including the number of hidden layers, the neuron count within each layer, the optimizer, and the learning rate. The number of hidden layers was chosen using a trial-and-error method, where several combinations were investigated, and the best-performing one was selected. The resulting model comprises 15 dense layers, with its architectural representation shown in Figure 6. The training phase encompassed 1,000 epochs, with a consistent batch size of 32 samples.

For the second output variable this study applies another DL model, with similar architecture and setting. The second DL model to predict average pore volume uses six dense layers, and the last dense layer utilizes a linear activation function. The optimizer function used was Adam. The model was trained for one thousand epochs, utilizing a batch size of 32 samples.

Using two separate DL models for the two output variables, despite having the same input, is chosen due to the distinct correlations each output has with different input variables, as indicated by the correlation heat map Figure 4. This allows each model to better capture the specific relationships relevant to its respective output. Additionally, separate models can be optimized with different architectures, avoiding the interference that might occur if a single model were used for both outputs. This approach also enhances performance and accuracy by tailoring the model parameters to each specific output. Also, it simplifies scalability and maintenance, allowing for independent updates and retraining as new data.

# 4 Results and discussion

The learning curve shown in Figure 7 represents the DL model's performance throughout the training and validation epochs. This curve illustrates the dynamic progress of the model's ability to make accurate predictions as it learns from the training data and generalizes to unseen data. The decreasing orange line in Figure 7 indicates that the model successfully reduces errors between the predicted and actual welding depths during training. This trend shows that the model effectively captures underlying patterns in the training data, leading to accurate predictions on new, unseen data during the validation phase.

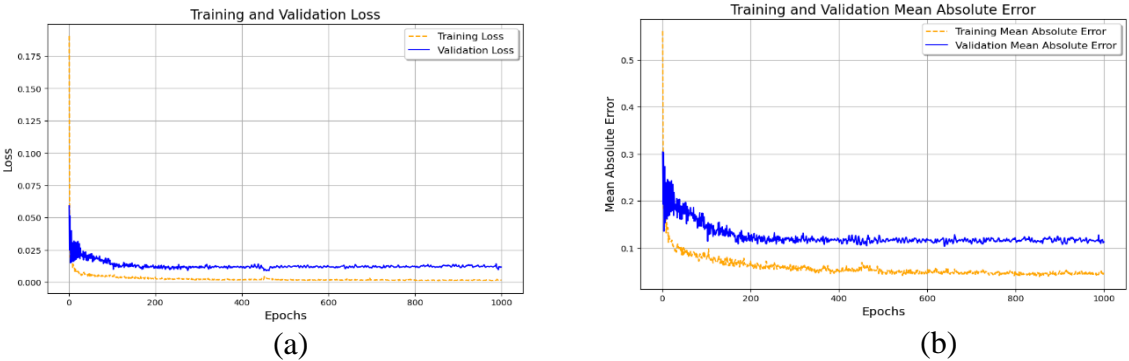

(a)　　　　　　　　　　　　　　　　　(b)

*Figure 7 displays the learning curve of the welding depth prediction network represented by two lines: curves illustrate loss values over 1000 epochs orange for the training phase and blue for the validation phase.*

A loss value of 0.0111 demonstrates that the model's predicted welding depth closely matches the actual values. The low MAE of 0.1079 further confirms that the model's predictions have a minimal average absolute difference from the actual values. These results indicate that the model effectively minimizes errors and provides accurate predictions for welding depth.

Figure 8 shows a small error between the model's predictions and the actual welding depths in the real dataset. The model's strong predictive capability is evident from this prediction, highlighting the model's reliability and accuracy.

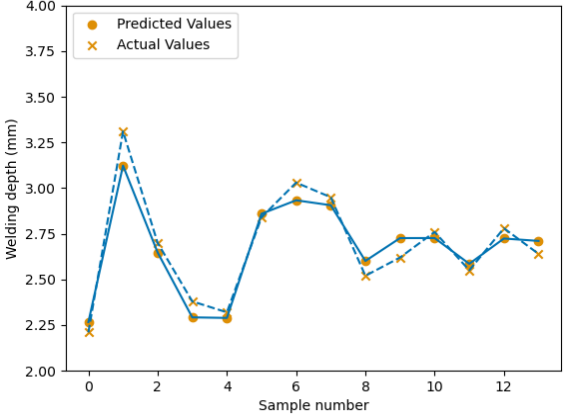

*Figure 8 shows the predicted welding depth in the test dataset, compared with the actual values.*

In contrast, Figure 9 illustrates the learning curves for predicting average pore volume, with a validation loss of 0.0067 and MAE of 0.0641. The lower position of the validation curve compared to the training curve shows potential underfitting. This underfitting is reflected in the $R^2$ value of 0.2095, indicating the model's limited ability to explain variance in the validation data. This result underscores the challenges in accurately predicting pore volume, as discussed in the introduction. Pore formation is a complex phenomenon influenced by various factors, and different types of pores have distinct input variables that affect the results.

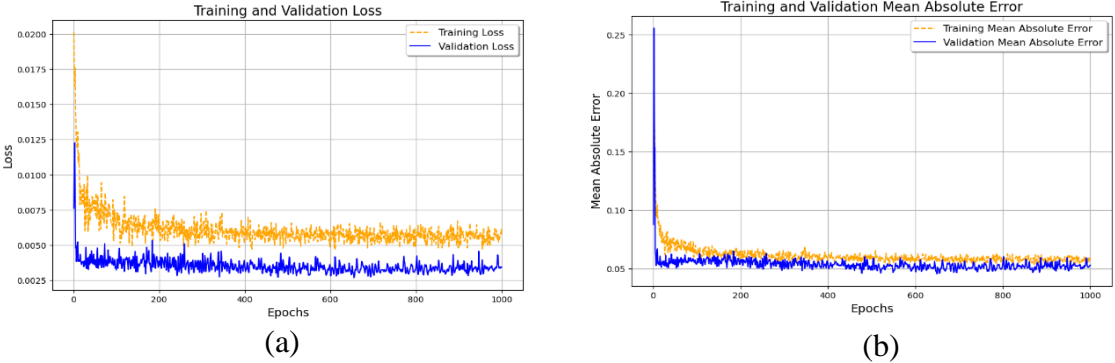

(a)  (b)

*Figure 9 displays the learning curve represented by two lines: orange for the training phase and blue for the validation phase. In (a), it shows loss values over 1000 epochs, while (b) illustrates MAE values over the same epochs. The lower position of the validation curve compared to the training curve*

The comparison of predicted and actual values in the test dataset, shown in Figure 10, further illustrates these challenges. Predicting pore volume is inherently difficult due to the complexity and variability of pore formation processes. Each type of pore has its own influencing factors, complicating the model's task.

Therefore, while the model demonstrates strong performance in predicting welding depths, it faces challenges in accurately capturing the variance in pore volume predictions. As shown in Figure 5(b), pore volume exhibits significant variations even for the same input values.

These findings suggest the need for more investigation or additional data to improve predictions for complex phenomena like pore formation. Future research should explore these paths to enhance model accuracy and applicability.

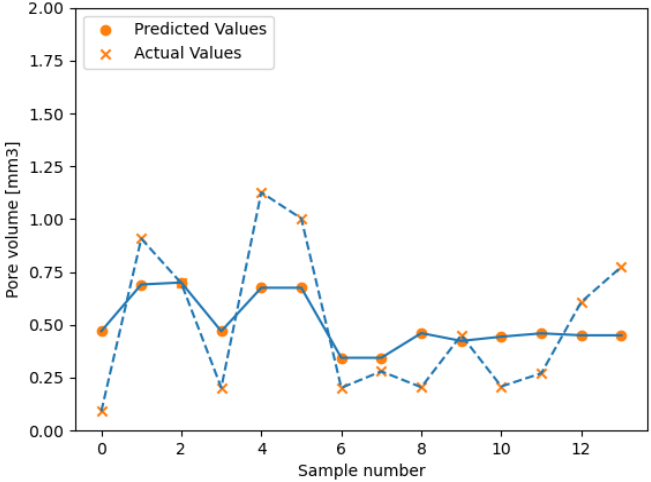

*Figure 10 predicted pore volume in the test dataset.*

# 5 Conclusion

In conclusion, this study presents a novel DL model developed to predict two critical welding KPCs: welding depth and average pore volume. This achievement integrates both numerical data analysis and extensive experimental validation, forming a robust foundation for this research. The DL model considers a comprehensive array of KICs, including geometric factors, path geometries, repetitions, feed rates, BLW ratio, machine energy output, and welding nozzle specifications, all crucial in determining weld quality.

The model effectively captures complex interactions among these input variables, resulting in high fitting accuracy. Statistical analysis of the small numerical dataset was essential, revealing patterns such as correlation strengths and output variable distributions to enhance model reliability.

Traditionally, the Design of Experiments (**DoEs**) optimizes welding parameters for desired weld geometries using such datasets, often complemented by simulations. However, our approach significantly accelerates this process, reducing prediction times from hours to seconds compared to simulation-based methods. This efficiency highlights the practical advantage of our approach.

Nevertheless, the DL model's limitations, common in laser welding applications, stem from its specificity to the hairpins made of copper dataset used in training. While comparing results with studies utilizing similar datasets [4-7, 38, 41] is feasible, differing data types pose challenges to direct comparisons.

Future research should validate the universality of our approach by inverse the DL model to predict desired weld depths and conducting real experiments for validation. This ongoing investigation aims to combine additional data sources: generated, simulation, and sensor data to further enhance weld quality and improve the model's accuracy and reliability.

In conclusion, this study lays a strong foundation for future research endeavors. The ultimate goal is to develop a robust online quality assurance system that leverages DL to predict KPCs and autonomously optimize KICs in real-time welding processes, meeting industry demands for enhanced efficiency and quality control.

# 6 Conflict of Interests

The authors declare no conflicts of interest.

# 7 Acknowledgment

Vinnova is greatly appreciated for funding this work through Production 2030 programme, grant number 2021-03693. Formas and the Swedish Energy Agency are appreciated for their valuable support and collaboration. Volvo Cars as a valued partner to our research works, is sincerely appreciated, for their continues support and materials.